\documentclass[11pt]{article}

\usepackage[preprint]{acl}

\usepackage{times}
\usepackage{latexsym}
\usepackage{amsmath}
\usepackage{amssymb} 
\usepackage{amsthm} 
\usepackage{mathtools}
\usepackage{bm}
\usepackage[most]{tcolorbox}
\usepackage{listings}
\usepackage{algorithm}
\usepackage{algpseudocode}
\usepackage{subcaption}
\usepackage{float}
\usepackage{booktabs} 
\usepackage{multirow}
\usepackage{makecell}       
\usepackage{array}          
\usepackage{tabularx}       
\usepackage{threeparttable} 
\usepackage{adjustbox}      
\usepackage{enumitem}
\usepackage{balance}
\usepackage{booktabs}
\usepackage{graphicx}
\usepackage{colortbl}
\usepackage{xcolor}
\usepackage{diagbox}
\definecolor{ourgray}{gray}{0.92}

\usepackage[T1]{fontenc}

\usepackage[utf8]{inputenc}

\usepackage{microtype}

\usepackage{inconsolata}

\usepackage{graphicx}

\title{Should Missing Modalities Always Be Necessary to Repair for Multi-modal Sentiment Analysis?}

\author{
  \textbf{Yubo Gao\textsuperscript{1,2}}%
  \thanks{These authors contributed equally to this work.},
  \textbf{Haotian Wu\textsuperscript{3}}\footnotemark[1],
  \textbf{Xiaoyu Xu\textsuperscript{5}}\footnotemark[1],
  \textbf{Yibo Yan\textsuperscript{1,2}},
\\
  \textbf{Hong Chen\textsuperscript{1,2}},
  \textbf{Ruoshui Peng\textsuperscript{1,2}},
  \textbf{Fei Pan\textsuperscript{5}},
  \textbf{Puay Siew Tan\textsuperscript{4}},
\\
  \textbf{Zhuoran Gao\textsuperscript{1,2}},
  \textbf{Yonghua Hei\textsuperscript{1,2}},
  \textbf{Jie Zhang\textsuperscript{3}},
  \textbf{Xuming Hu\textsuperscript{1,2}}%
  \thanks{Corresponding author.}
\\[0.45em]
  \textsuperscript{1}The Hong Kong University of Science and Technology (Guangzhou),\\
  \textsuperscript{2}The Hong Kong University of Science and Technology,\\
  \textsuperscript{3}Nanyang Technological University,\\
  \textsuperscript{4}Singapore Institute of Manufacturing Technology, A*STAR,\\
  \textsuperscript{5}Lingnan University
\\[0.35em]
  \small{
    \textbf{Emails:}
    \href{mailto:ygao704@connect.hkust-gz.edu.cn}
    {ygao704@connect.hkust-gz.edu.cn},
    \href{mailto:xuminghu@hkust-gz.edu.cn}
    {xuminghu@hkust-gz.edu.cn}
  }
}

\begin{document}
\maketitle
\begin{abstract}
Existing methods for multimodal sentiment analysis (MSA) under missing modalities usually follow a repair-first paradigm. We revisit this assumption and ask: \emph{should every missing modality be repaired?} A per-sample oracle analysis shows the answer is not always: full-modality input is optimal for only a small fraction of samples, and every modality subset is preferred by some samples. These results suggest that adding or repairing modalities may not always improve prediction, and that the utility of each modality is sample-dependent. Building on this finding, we propose \textbf{S}ufficiency-\textbf{I}nformed \textbf{E}vidential \textbf{V}al\textbf{vE} (\textbf{SIEVE}) that turns ``whether to repair'' into an explicit, learnable decision at the sample level. SIEVE compares a direct prediction branch with a repair branch, derives an empirical sufficiency signal from their per-sample loss gap, and routes each input through an evidential gate that jointly models sufficiency and its epistemic uncertainty. SIEVE is repair-agnostic: it operates as a plug-and-play decision on top of any explicit or implicit repair module, without modifying its internal design. Experiments on CMU-MOSI and IEMOCAP show that SIEVE consistently improves representative repair backbones across evaluated missing rates, and approaches the per-sample dual-branch achievable optimum. 
\end{abstract}

\section{Introduction}
Multimodal sentiment analysis (MSA) aims to infer human affect by integrating language, acoustic, and visual signals~\cite{zadeh2016mosi,zadeh2018multimodal}. Prior work has developed diverse fusion and representation learning strategies to leverage the complementarity of these cues~\cite{zadeh2017tensor,tsai2019multimodal,hazarika2020misa,yu2021learning}. However, most canonical methods rest on a full-modality assumption that is fragile in real-world deployments: acoustic signals may be corrupted by noise, visual cues may be occluded or withheld for privacy, and transcripts may be degraded by recognition errors or access restrictions~\cite{zhao2021missing,lian2023gcnet}. These incomplete observations have motivated a growing body of work on robust MSA under missing modalities.
\vspace{-0.2cm}
\begin{figure}[H]
    \centering
    \includegraphics[width=1.0\linewidth]{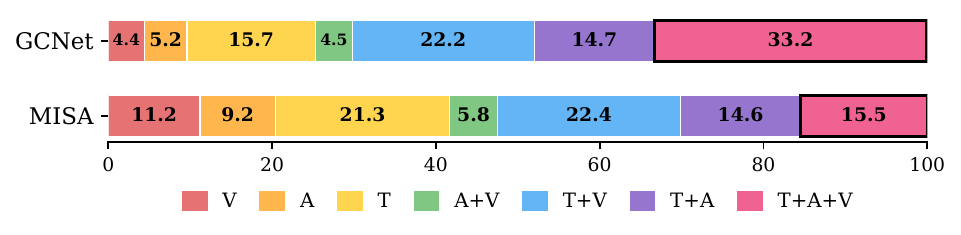}
    \vspace{-0.4cm}
    \caption{Best-performing modality combination per sample on CMU-MOSI (\%).}
    \label{fig:oracle}
\end{figure}
\vspace{-0.4cm}

Existing robust MSA approaches largely adopt a repair-oriented view of missing modalities, either through \emph{explicit repair} that reconstructs absent modalities from the observed ones~\cite{pham2019found,zhao2021missing,lian2023gcnet,guo2024multimodal,wang2023incomplete,wang2023distribution}, or through \emph{implicit repair} that compensates for the resulting semantic deficiency at the representation or decision level~\cite{wei2023mmanet,li2024unified,li2024toward,li2024correlation,zhuang2025cmad,zhang2024towards,zhuang2025hyper,xu2024leveraging,zhuang2026tmdc}. Despite their different mechanisms, these methods share a repair-first tendency: whenever a modality is absent, they intervene before prediction, without examining whether such repair is necessary for the current sample.
In this paper, we revisit this assumption and ask: \textit{Should every missing modality be repaired?} A diagnostic oracle analysis on CMU-MOSI (Figure~\ref{fig:oracle}) suggests otherwise. We evaluate two representative MSA models, MISA and GCNet, under each of the seven possible modality combinations on every test sample, and identify the combination that yields the lowest prediction error. The result is striking: full-modality input is the best choice for only 15.5\% of samples under MISA and 33.2\% under GCNet, while every one of the seven subsets is preferred for some samples.
This pattern holds across both models and reveals a more fundamental principle: the utility of a modality is sample-dependent. For some samples, an additional modality introduces redundancy or noise instead of complementary evidence, leaving the model better off without it. Thus, reconstructing or compensating for every missing modality may be unnecessary, and in some cases may even hurt prediction.
This reshapes the central question of missing-modality MSA. Instead of asking only \textit{how to repair a missing modality}, a model should first decide \textit{whether repair is needed for the current sample}. When the observed modalities already carry enough evidence, direct prediction may be preferable; when they do not, repair remains beneficial. Yet existing methods invoke repair whenever any modality is absent, leaving this sample-level decision unaddressed.

To this end, we propose \textbf{S}ufficiency-\textbf{I}nformed \textbf{E}vidential \textbf{V}alv\textbf{E} (\textbf{SIEVE}) for selective modality repair. SIEVE turns 
``whether to repair'' into an explicit, learnable decision at the sample level. For each input, SIEVE compares two prediction pathways: a \emph{direct branch} that predicts from the observed modalities, and a \emph{repair branch} that first invokes a repair module and then predicts. An evidential sufficiency valve then routes the sample between the two pathways. SIEVE is a plug-and-play, repair-agnostic framework compatible with both explicit and implicit repair methods.
The central challenge is that no external ground-truth signal exists for whether a sample's observed modalities are sufficient. SIEVE addresses this by deriving an empirical sufficiency signal directly from the per-sample loss gap between the two branches, which reveals whether direct prediction is at least as accurate as repaired prediction. To make routing reliable under noisy signals, SIEVE further models sufficiency as an evidential distribution, jointly capturing the sufficiency estimate and its epistemic uncertainty to enable uncertainty-aware routing.
Our contributions are threefold:
\vspace{-0.2cm}
\begin{itemize}[leftmargin=*]
\item We challenge the repair-first assumption: a per-sample oracle analysis shows that forcing repair on every incomplete input may not be helpful for a substantial fraction of samples.
\vspace{-0.1cm}
\item We propose \textbf{SIEVE}, a selective repair framework that learns a per-sample decision of \emph{whether to repair} from the dual-branch loss gap, without external sufficiency labels.
\vspace{-0.1cm}
\item As a plug-and-play decision framework, SIEVE consistently improves diverse repair backbones across benchmarks and missing rates.
\end{itemize}
\section{Related Work}

\noindent\textbf{MSA with Missing Modalities.}
Multimodal sentiment analysis (MSA) integrates language, acoustic, and visual signals to infer human affect. Since real-world inputs are often incomplete, recent studies have explored robust MSA under missing modalities, which can be broadly grouped into explicit and implicit modality repair (see Appendix~\ref{app:related-work} for a detailed review).
\noindent\textit{Explicit modality repair} reconstructs, imagines, or approximates absent modalities from the observed ones before prediction~\cite{pham2019found,zhao2021missing,lian2023gcnet,guo2024multimodal,wang2023distribution,wang2023incomplete}.
\noindent\textit{Implicit modality repair} compensates for the resulting semantic deficiency at the representation or decision level, either by distilling full-modality knowledge into incomplete-modality models~\cite{wei2023mmanet,li2024unified,li2024toward,li2024correlation,zhuang2025cmad}, or by intrinsic evidence compensation without a full-modality teacher~\cite{zhang2024towards,zhuang2025hyper,xu2024leveraging,zhuang2026tmdc}.
Overall, existing methods study \emph{how} to repair missing modalities. In contrast, SIEVE asks \emph{whether} repair is needed for each sample, enabling selective repair when the observed modalities are insufficient and direct prediction when they already suffice.

\noindent\textbf{Subjective Logic and Evidential Deep Learning}
Subjective Logic~\cite{josang2001logic,josang2016subjective} represents uncertainty through subjective opinions.
For a binary proposition, an opinion consists of belief $b$, disbelief $d$, uncertainty $u$, and a base rate $a$, where $b+d+u=1$; the base rate specifies how the uncommitted uncertainty mass is projected to a probability.
Under the conjugate Beta view, the parameters $\alpha$ and $\beta$ encode evidence for the two outcomes, and a smaller total concentration $\alpha+\beta$ corresponds to higher epistemic uncertainty.
Evidential deep learning~\cite{xu2024reliable,zhuang2025hyper} brings this principle into neural networks by directly predicting the parameters of an appropriate evidence distribution, such as Dirichlet distributions (and their binary special case, Beta)~\cite{han2021trusted,han2022trusted} for classification~\cite{sensoy2018evidential} and Normal-Inverse-Gamma distributions for regression~\cite{amini2020deep}.

\section{Method}
\subsection{Overview and Problem Setup}
Many existing missing-modality MSA methods follow a \textit{repair-first} assumption: whenever a modality is unavailable, the model should reconstruct, distill, or otherwise compensate for it before prediction. As our pilot study in Section~1 shows, however, the utility of a modality is sample-dependent, and forcing repair on every incomplete sample may be harmful. This shifts the central question from \textit{how to repair missing modalities} to \textit{whether repair is needed for the current sample}.
\begin{figure*}[htp]
    \centering
    \includegraphics[width=1.0\linewidth]{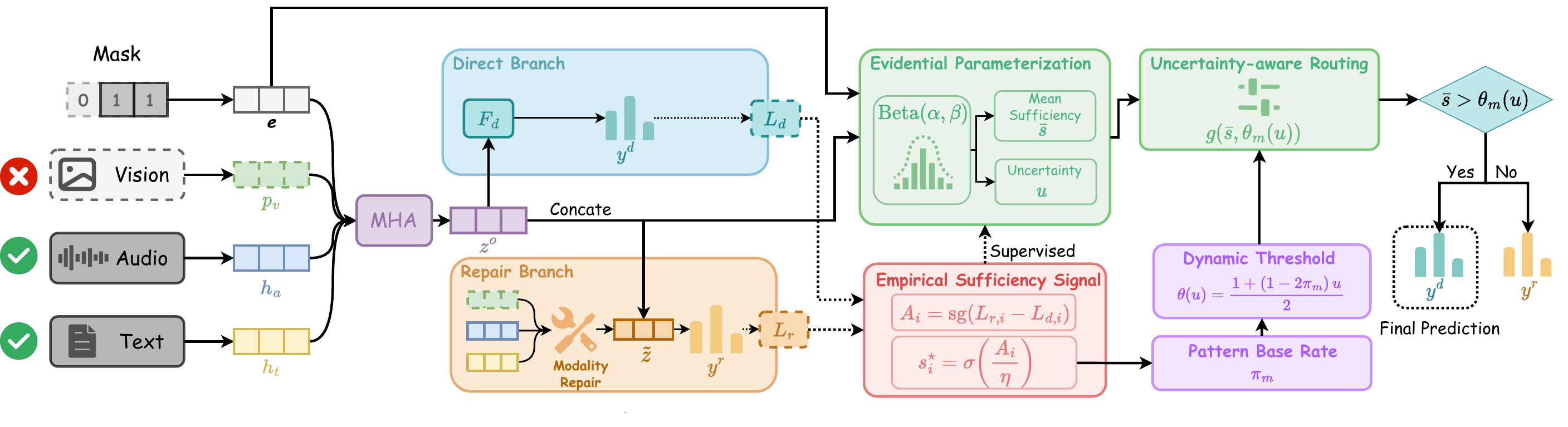}
    \vspace{-0.6cm}
    \caption{Overview of SIEVE. Direct and repair branch yield two predictions, whose per-sample loss gap supervises parameterization of the evidential valve. The valve routes each sample through an uncertainty-aware threshold.}
\label{fig:framework}
\end{figure*}

\paragraph{Framework overview.}
SIEVE operationalizes selective modality repair through a \textbf{sample-adaptive sufficiency assessment framework} (Figure~\ref{fig:framework}). For each incomplete input, SIEVE constructs two comparable prediction pathways. The first is a \textit{direct branch}, which predicts from the observed modalities only. The second is a \textit{repair branch}, which first applies a repair module $\mathcal{R}_\omega$ and then performs prediction. The repair module is used as an interchangeable component, so $\mathcal{R}_\omega$ can be instantiated by either explicit or implicit modality methods. In this way, SIEVE does not replace existing repair models, but adds a decision module that determines whether their intervention is needed for the current input.

The central difficulty is that sufficiency is not directly annotated and cannot be inferred from the missing pattern alone. Two samples with the same missing modalities may require different treatments, because the observed modalities may already contain enough task-relevant evidence for one sample but remain inadequate for another. SIEVE therefore defines sufficiency operationally within the dual-branch system: the observed modalities are treated as empirically sufficient when direct prediction is no worse than repair-based prediction. This yields an \textit{empirical sufficiency signal} from the per-sample loss gap between the two branches, which provides supervision for an \textit{evidential sufficiency valve}. The valve formulates sufficiency as a binary Subjective Logic proposition, namely whether the observed modalities provide enough evidence to skip repair, and predicts a Beta distribution over this proposition. Evidence for sufficiency favors the direct branch, evidence for insufficiency favors the repair branch, and high uncertainty defers to the missing-pattern prior, falling back to whichever pathway is more likely to succeed under the current missing pattern.

\paragraph{Problem setup.}
Let $\mathcal{M}=\{l,a,v\}$ denote the language, acoustic, and visual modalities. For the $i$-th sample, modality availability is encoded by a binary mask $\mathbf{m}_i\in\{0,1\}^{|\mathcal{M}|}$, where $m_i^k=1$ if modality $k$ is observed and $0$ otherwise. We write $\mathcal{O}_i=\{k\in\mathcal{M}:m_i^k=1\}$ for the observed modality set and assume $|\mathcal{O}_i|\geq 1$. Each instance is $(\{\mathbf{x}_i^k\}_{k\in\mathcal{O}_i},\mathbf{m}_i,y_i)$, with $\mathbf{x}_i^k$ the input feature of modality $k$ and $y_i$ the sentiment label. The goal is to predict $y_i$ from the observed modalities while jointly deciding, on a per-sample basis, whether to activate the repair module.

\subsection{Dual-Branch Prediction}

This module builds the two prediction pathways used by SIEVE. The direct branch predicts from the observed modalities alone, while the repair branch first applies the repair module and then predicts. Their per-sample loss gap will later supervise the sufficiency valve (Section~\ref{sec:valve}).
\vspace{-0.2cm}
\paragraph{Observed representation.}
We first encode the observed modalities into a single feature vector $\mathbf{z}_i^{o}$ that both branches share. For each modality, SIEVE uses its encoder if the modality is observed, and a learnable placeholder otherwise:
\vspace{-0.1cm}
\begin{equation}
\bar{\mathbf{h}}_i^k=
\begin{cases}
E_k(\mathbf{x}_i^k), & m_i^k=1,\\
\mathbf{p}^k, & m_i^k=0,
\end{cases}
\qquad k\in\mathcal{M},
\end{equation}
\vspace{-0.1cm}
where $E_k(\cdot)$ is the modality-specific encoder, $\mathbf{p}^k\in\mathbb{R}^{d}$ is a learnable placeholder, and $\bar{\mathbf{h}}_i^k\in\mathbb{R}^{d}$ is the resulting modality representation. The placeholder only signals absence; it does not synthesize content.
We also embed the binary mask into a continuous vector:
\vspace{-0.4cm}
\begin{equation}
\mathbf{e}_i=W_m\mathbf{m}_i+\mathbf{b}_m,
\end{equation}
with trainable parameters $W_m\in\mathbb{R}^{d\times|\mathcal{M}|}$ and $\mathbf{b}_m\in\mathbb{R}^{d}$. Stacking the modality representations into a sequence $\mathbf{H}_i=[\bar{\mathbf{h}}_i^l;\bar{\mathbf{h}}_i^a;\bar{\mathbf{h}}_i^v]$ and using $\mathbf{e}_i$ as the query, we aggregate them via multi-head attention:
\vspace{-0.2cm}
\begin{equation}
\mathbf{z}_i^{o}=\operatorname{Pooling}(\operatorname{MHA}(\mathbf{e}_i,\mathbf{H}_i,\mathbf{H}_i)),
\end{equation}
where $\operatorname{MHA}(\cdot,\cdot,\cdot)$ denotes multi-head attention with query $Q$, key $K$, and value $V$. The output $\mathbf{z}_i^{o}\in\mathbb{R}^{d}$ therefore summarizes the evidence currently available.

\paragraph{Direct and repair branches.}
The direct branch predicts from $\mathbf{z}_i^{o}$ alone:
\vspace{-0.2cm}
\begin{equation}
\hat{y}_i^{d}=F_d(\mathbf{z}_i^{o}),
\end{equation}
where $F_d(\cdot)$ is the direct-branch predictor.

The repair branch first applies the repair module $\mathcal{R}_{\omega}$ to obtain a repaired representation:
\vspace{-0.2cm}
\begin{equation}
\tilde{\mathbf{z}}_i=\mathcal{R}_{\omega}(\mathbf{H}_i,\mathbf{m}_i),
\end{equation}
where $ \mathcal{R}_\omega$ can be any explicit or implicit modality repair method. For methods without an explicit repaired modality, we use their compensated representation as $\tilde{\mathbf{z}}_i$. The repair branch then concatenates the observed and repaired representations and feeds them to the repair-branch predictor:
\begin{equation}
\hat{y}_i^{r}=F_r([\mathbf{z}_i^{o};\tilde{\mathbf{z}}_i]),
\end{equation}
where $[\cdot;\cdot]$ denotes concatenation and $F_r(\cdot)$ is the repair-branch predictor.

\subsection{Evidential Sufficiency Valve}
\label{sec:valve}

The valve decides, for each sample, whether the observed modalities are sufficient for direct prediction. Since no ground-truth label exists for this decision, we define sufficiency operationally with respect to the dual-branch system: the observed modalities are \textit{empirically sufficient} if direct prediction is no worse than repair-based prediction. This turns sufficiency into an observable quantity: comparable per-sample losses of the two branches, and removes the need for any external supervision.

\paragraph{Empirical sufficiency signal.}
We compute the per-sample task losses of the two branches,
\vspace{-0.2cm}
\begin{equation}
L_{d,i}=\ell(\hat{y}_i^{d},y_i),
\
L_{r,i}=\ell(\hat{y}_i^{r},y_i),
\end{equation}
where $\ell(\cdot,\cdot)$ is the task loss. The empirical advantage of direct prediction is the loss gap
\vspace{-0.2cm}
\begin{equation}
A_i=\operatorname{sg}(L_{r,i}-L_{d,i}),
\end{equation}
where $\operatorname{sg}(\cdot)$ is stop-gradient: $A_i>0$ means the direct branch is better, and $A_i<0$ means repair is helpful. We convert this gap into a soft sufficiency target via a temperature-scaled sigmoid:
\vspace{-0.2cm}
\begin{equation}
s_i^{\star}=\sigma\!\left(A_i / \eta\right),
\end{equation}
with $\eta>0$. Thus $s_i^{\star}$ approaches $1$ when direct prediction wins and $0$ when repair wins. Here, $s_i^{\star}$ serves as a soft internal supervised anchor (empirical sufficiency targets) for training the valve. It is used to fit the predicted sufficiency estimate $\bar{s}_i$.

\paragraph{Evidential parameterization.}
The empirical sufficiency target $s_i^{\star}$ is inherently noisy, since it is derived from the per-sample losses of two branches that co-evolve during training. The valve should therefore predict not only a sufficiency score, but also its uncertainty. To this end, we parameterize a Beta distribution over the sufficiency probability of sample $i$, with its parameters predicted from the observed evidence.
The valve first maps the observed representation $\mathbf{z}_i^{o}$ and the mask embedding $\mathbf{e}_i$ to a hidden state, and then predicts the Beta parameters
\vspace{-0.2cm}
\begin{equation}
\begin{aligned}
\mathbf{v}_i&=\operatorname{ReLU}\!\left(W_1[\mathbf{z}_i^{o};\mathbf{e}_i]+\mathbf{b}_1\right),\\
\alpha_i&=\operatorname{Softplus}(\mathbf{w}_{\alpha}^{\top}\mathbf{v}_i+c_{\alpha})+1,\\
\beta_i&=\operatorname{Softplus}(\mathbf{w}_{\beta}^{\top}\mathbf{v}_i+c_{\beta})+1,
\end{aligned}
\end{equation}
where the $+1$ shift ensures $\alpha_i,\beta_i>1$ and stabilizes training. Intuitively, $\alpha_i$ accumulates evidence for sufficiency, and $\beta_i$ accumulates evidence for insufficiency. This induces the Beta posterior:
\vspace{-0.2cm}
\begin{equation}
p(q_i \mid \mathbf{z}_i^o, \mathbf{e}_i)=\mathrm{Beta}(q_i;\alpha_i,\beta_i),
\end{equation}
where $q_i\in[0,1]$ denotes the latent sufficiency probability. We use the posterior mean as the predicted sufficiency score:
\vspace{-0.2cm}
\begin{equation}
\bar{s}_i=\frac{\alpha_i}{\alpha_i+\beta_i}.
\end{equation}
\vspace{-0.1cm}
Following Subjective Logic~\citep{josang2016subjective,han2021trusted}, the same Beta distribution decomposes into belief ($b_i$), disbelief($d_i$), and uncertainty ($u_i$) masses,
\vspace{-0.2cm}
\begin{equation}
\small
b_i=\frac{\alpha_i-1}{\alpha_i+\beta_i},\
d_i=\frac{\beta_i-1}{\alpha_i+\beta_i},\
u_i=\frac{2}{\alpha_i+\beta_i},
\end{equation}
satisfying $b_i+d_i+u_i=1$. A smaller total evidence $\alpha_i+\beta_i$ yields larger $u_i$, so $u_i$ measures the valve's epistemic uncertainty.
\vspace{-0.2cm}
\paragraph{Uncertainty-aware routing.}
A reliable valve must adapt its routing threshold to both the per-sample uncertainty and the prior repair necessity of each missing pattern. To this end, we design a dynamic threshold that combines an evidential uncertainty term with a missing-pattern-conditioned base rate ($\pi_{m_i} \in [0, 1]$):
\begin{equation}
\label{eq:threshold}
\theta(u_i) = \frac{1 + (1 - 2\pi_{m_i})\, u_i}{2},
\end{equation}
where $\pi_{m_i} \in [0, 1]$ estimates the empirical probability of direct-branch sufficiency under the missing pattern $m_i$. We maintain $\pi_m$ online as an exponential moving average over the empirical sufficiency targets within each missing pattern:
\begin{equation}
\pi_{m} \leftarrow (1 - \kappa)\, \pi_{m} + \kappa \cdot \frac{1}{|\mathcal{B}_m|} \sum\nolimits_{i \in \mathcal{B}_m} s_i^\star,
\end{equation}
where $\mathcal{B}_m$ collects the samples in the current batch with missing pattern $m$, and $\kappa\in(0, 1]$ is the update rate. Intuitively, when a missing pattern frequently yields sufficient direct prediction, $\pi_m$ grows and the threshold drops, allowing the valve to more readily skip repair. When repair is typically needed, $\pi_m$ shrinks and the threshold rises, biasing routing toward repair. The dynamic threshold thus calibrates itself to the empirical repair necessity of each missing pattern, all without supervision (Appendix~\ref{app:threshold}).
At inference, the routing decision is hard (without Gumbel noise):
\begin{equation}
g_i^{\mathrm{hard}}=\mathbb{I}\!\left[\bar{s}_i>\theta(u_i)\right],
\end{equation}
with $g_i^{\mathrm{hard}}=1$ routing to the direct branch and $g_i^{\mathrm{hard}}=0$ to the repair branch.

At training, hard routing blocks gradients. We therefore use a Gumbel-Sigmoid relaxation~\cite{jang2016categorical} around the same decision boundary. Define the routing margin:
\begin{equation}
r_i=\operatorname{logit}(\bar{s}_i)-\operatorname{logit}(\theta(u_i)),
\end{equation}
with $\operatorname{logit}(x)=\log\frac{x}{1-x}$, so that $r_i>0$ is equivalent to $\bar{s}_i>\theta(u_i)$. The differentiable soft routing weight is
\begin{equation}
\label{eq:gumbel}
\small
g_i=\sigma\!\left(\frac{r_i+\log\rho_i-\log(1-\rho_i)}{\tau}\right),
\ \rho_i\sim\operatorname{Uniform}(0,1),
\end{equation}
where $\tau>0$ is the relaxation temperature. During training $g_i\in(0,1)$ softly mixes the two branches, and as $\tau$ decreases $g_i$ approaches a near-binary stochastic routing variable.

\subsection{Optimization and Inference}

Training has two goals: stable prediction branches and a reliable sufficiency valve. Because the empirical sufficiency target $s_i^{\star}$ depends on the current accuracy of both branches, optimizing the valve too early uses unreliable supervision. We therefore adopt a two-stage protocol.

\paragraph{Stage 1: Warmup.}
We train the two branches with equal weights, keeping the valve frozen:
\begin{equation}
\small
\mathcal{L}_{\mathrm{warm}}
=
\frac{1}{N}\sum_{i=1}^{N}\frac{1}{2}\!\left[\ell(\hat{y}_i^{d},y_i)+\ell(\hat{y}_i^{r},y_i)\right]
+\lambda_R\,\mathcal{L}_{R},
\end{equation}
where $N$ is the number of training samples, $\mathcal{L}_R$ is the optional internal loss of the repair module, and $\lambda_R$ is its weight. This stage stabilizes both branches so that the loss gap $A_i$ becomes a meaningful supervision signal for the valve.

\paragraph{Stage 2: Joint training.}
We then jointly optimize the branches and the valve with three complementary losses.
The \textit{routed prediction loss} weights the two branches by the soft routing weight $g_i$ from Eq.~(\ref{eq:gumbel}):
\vspace{-0.2cm}
\begin{equation}
\small
\mathcal{L}_{\mathrm{pred}}
=
\frac{1}{N}\sum_{i=1}^{N}\!\left[g_i\,\ell(\hat{y}_i^{d},y_i)+(1-g_i)\,\ell(\hat{y}_i^{r},y_i)\right].
\end{equation}

The \textit{valve calibration loss} aligns the predicted sufficiency distribution with the empirical sufficiency target $s_i^{\star}$. Specifically, the valve predicts a Beta distribution parameterized by $(\alpha_i,\beta_i)$, and we maximize the density assigned to the $s_i^{\star}$:
\vspace{-0.2cm}
\begin{equation}
\small
\begin{split}
    \mathcal{L}_{\mathrm{valve}}
=
-\frac{1}{N}\sum_{i=1}^{N}
\Big[
(&\alpha_i-1)\log \tilde{s}_i^{\star}
+
(\beta_i-1)\log(1-\tilde{s}_i^{\star})
 \\
&-\log B(\alpha_i,\beta_i)
\Big],
\end{split}
\end{equation}
\vspace{-0.1cm}
where $B(\alpha_i,\beta_i)$ is the Beta function. We use $\tilde{s}_i^{\star}=\operatorname{clip}(s_i^{\star},\epsilon,1-\epsilon)$ to keep the $s_i^{\star}$ away from the boundary values $0$ and $1$, 
where the Beta log-density contains unstable logarithmic terms.

The \textit{evidence regularizer} penalizes evidence that contradicts the empirical target. Inspired by the soft-label adaptation of evidential deep learning~\citep{sensoy2018evidential,han2021trusted}, we form adjusted Beta parameters that mask out the target-aligned evidence and retain only the wrong-direction component:
\begin{equation}
\tilde{\alpha}_i=s_i^{\star}+(1-s_i^{\star})\alpha_i,\
\tilde{\beta}_i=(1-s_i^{\star})+s_i^{\star}\beta_i.
\end{equation}
When $s_i^{\star}\!\to\!1$ (direct branch better), $\tilde{\alpha}_i\!\to\!1$ while $\tilde{\beta}_i$ retains $\beta_i$; the converse holds when $s_i^{\star}\!\to\!0$. Pulling the adjusted Beta toward the non-informative prior therefore penalizes only evidence assigned in the wrong direction:
\begin{equation}
\small
\mathcal{L}_{\mathrm{ev}}
=
\frac{1}{N}\sum_{i=1}^{N}D_{\mathrm{KL}}\!\left[\operatorname{Beta}(\tilde{\alpha}_i,\tilde{\beta}_i)\,\|\,\operatorname{Beta}(1,1)\right].
\end{equation}

The overall objective is
\begin{equation}
\small
\mathcal{L}
=
\mathcal{L}_{\mathrm{pred}}
+\lambda_v\,\mathcal{L}_{\mathrm{valve}}
+\lambda_e\,q(t)\,\mathcal{L}_{\mathrm{ev}}
+\lambda_R\,\mathcal{L}_{R},
\end{equation}
where $\lambda_v,\lambda_e,\lambda_R$ are balancing coefficients and $q(t)=\min(t/T_e,1)$ ramps up the evidence regularizer over the first $T_e$ epochs ($t$ denotes the current epoch). We also anneal the Gumbel-Sigmoid temperature from $\tau_0$ to $\tau_{\min}$ during this stage.

\paragraph{Inference.}
At test time, SIEVE computes $\bar{s}_i$ and $u_i$ from the observed modalities and the mask, and applies the hard valve:
\begin{equation}
\small
\hat{y}_i=
\begin{cases}
F_d(\mathbf{z}_i^{o}), & \bar{s}_i>\theta(u_i),\\[2pt]
F_r([\mathbf{z}_i^{o};\mathcal{R}_{\omega}(\mathbf{H}_i,\mathbf{m}_i)]), & \bar{s}_i\leq\theta(u_i).
\end{cases}
\end{equation}
Repair is invoked when the valve estimates the observed modalities to be insufficient under the uncertainty- and pattern-adjusted threshold. This selective mechanism retains the benefit of repair when needed, while avoiding unnecessary intervention when the observed modalities suffice.
\section{Experiments}

\subsection{Experimental Setup}
\label{sec:exp_setup}

\noindent\textbf{Datasets.}
We conduct experiments on two standard MSA benchmarks.
\textbf{CMU-MOSI}~\citep{zadeh2016mosi} provides 2{,}199 opinion video clips labeled with continuous sentiment intensities in $[-3, +3]$, split into 1{,}284 / 229 / 686 utterances for training, validation, and testing.
\textbf{IEMOCAP}~\citep{busso2008iemocap} comprises 5{,}531 dyadic conversational utterances; following~\citet{lian2023gcnet}, we focus on the four-class subset (\emph{happy}, \emph{sad}, \emph{angry}, \emph{neutral}) and adopt leave-one-session-out 5-fold cross-validation protocol~\cite{lian2026merbench}.

\noindent\textbf{Implementation Details.}
Following prior work~\citep{lian2023gcnet,xu2024leveraging,zhuang2026tmdc}, we extract utterance-level features as follows:
wav2vec-large~\cite{schneider2019wav2vec} for audio, DeBERTa-large~\cite{he2020deberta} for text, and MTCNN followed by MA-Net~\cite{zhao2021learning} for vision~\cite{zhang2016joint}.
We follow the \textbf{random missing protocol} of GCNet~\citep{lian2023gcnet} so that no complete-modality supervision leaks into training.
The missing rate is $MR = 1 - \frac{\sum_i c_i}{3L}$, where $m_i \in \{1, 2, 3\}$ denotes the count of observed modalities for the $i$-th sample and $L$ is the dataset size.
For every sample, each modality is independently dropped with probability $MR$, subject to $c_i \geq 1$ (hence $MR \leq 2/3 \approx 0.67$; $MR = 0.7$ approximates this limit~\cite{lian2023gcnet}).
Identical missing patterns are sampled across training, validation, and testing, and the dropped features remain inaccessible throughout. Detailed hyperparameter settings are provided in Appendix~\ref{app:hyperparameters}.
All results are averaged over five random seeds on a single NVIDIA A100-PCIE-40GB.

\paragraph{Baselines and Evaluation Metrics.}
We compare SIEVE with two groups of baselines.
The first group consists of general MSA methods originally designed for the full-modality setting, including MISA~\citep{hazarika2020misa} and Self-MM~\citep{yu2021learning}.
The second group consists of missing-modality MSA methods.
\emph{Explicit modality repair} baselines include MMIN~\citep{zhao2021missing}, GCNet~\citep{lian2023gcnet}, DiCMoR~\citep{wang2023distribution}, and IMDer~\citep{wang2023incomplete}.
\emph{Implicit modality repair} baselines fall into two sub-families.
The distillation-based methods include CorrKD~\citep{li2024correlation} and CMAD~\citep{zhuang2025cmad};
the intrinsic evidence compensation methods include LNLN~\citep{zhang2024towards}, MoMKE~\citep{xu2024leveraging}, and TMDC~\citep{zhuang2026tmdc}.

For evaluation, we adopt dataset-appropriate metrics consistent with prior work~\cite{lian2023gcnet}.
On CMU-MOSI, where labels are continuous, we report binary accuracy (\textit{Acc-2}) and \textit{Weighted F1} computed over the positive/negative polarity.
On IEMOCAP, we report four-class accuracy (\textit{Acc-4}) and \textit{Weighted F1}.

\subsection{Experimental Results and Analysis}
\vspace{-0.2cm}
\begin{table}[htp]
\caption{Performance comparison under extreme missing modality ($MR=0.7$). Best results in \textbf{bold}, second-best \underline{underlined}. ``F1'' denotes Weighted F1.}
\vspace{-0.4cm}
\label{tab:main}
\begin{center}
\resizebox{\linewidth}{!}{%
\begin{tabular}{l|cc|cc}
\toprule
 & \multicolumn{2}{c|}{\textbf{CMU-MOSI}} & \multicolumn{2}{c}{\textbf{IEMOCAP}} \\
\cmidrule(lr){2-3} \cmidrule(lr){4-5}
\textbf{Method} & Acc-2 ($\uparrow$) & F1 ($\uparrow$) 
                & Acc-4 ($\uparrow$) & F1 ($\uparrow$) \\
\midrule
MISA           & 0.6128 & 0.5942 & 0.5138 & 0.5073 \\
Self-MM        & 0.5701 & 0.5433 & 0.5446 & 0.5443 \\
\midrule
MMIN           & 0.5655 & 0.5248 & 0.5531 & 0.5538 \\
IMDer          & 0.5741 & 0.5487 & 0.5893 & 0.5891 \\
DiCMoR         & 0.5915 & 0.5769 & 0.5572 & 0.5579 \\
CMAD           & 0.6250 & 0.6184 & 0.6026 & 0.6017 \\
TMDC           & \underline{0.7043} & 0.6891 & 0.6326 & 0.6298 \\
LNLN           & 0.6220 & 0.6242 & 0.5282 & 0.5192 \\
\midrule
GCNet          & 0.7027 & 0.6998 & \underline{0.6816} & \underline{0.6840} \\
\rowcolor{ourgray} \quad + SIEVE & \textbf{0.7165} & \textbf{0.7100} & \textbf{0.6942} & \textbf{0.6968} \\
CorrKD         & 0.5854 & 0.5642 & 0.5575 & 0.5570 \\
\rowcolor{ourgray} \quad + SIEVE & 0.6174 & 0.5993 & 0.5756 & 0.5759 \\
MoMKE            & 0.6705 & 0.6718 & 0.6437 & 0.6405 \\
\rowcolor{ourgray} \quad + SIEVE & 0.6982 & \underline{0.6999} & 0.6555 & 0.6545 \\
\bottomrule
\end{tabular}%
}
\end{center}
\end{table}
\vspace{-0.2cm}
\noindent\textbf{Comparison with Baselines.} Table~\ref{tab:main} compares SIEVE with representative general and robust MSA methods under the extreme missing setting ($MR = 0.7$).
To verify that selective repair is a general principle rather than a method-specific trick, we integrate SIEVE into three representative repair paradigms: GCNet (explicit graph-based reconstruction), CorrKD (distillation-based implicit repair), and MoMKE (intrinsic evidence compensation with modality experts). On both CMU-MOSI and IEMOCAP, SIEVE consistently improves all three backbones, and GCNet+SIEVE further attains the best overall performance among all compared methods. The improvements span all three repair paradigms, confirming that SIEVE complements existing repair models without altering their internal design: when the observed modalities are sufficient, SIEVE allows direct prediction; otherwise, it preserves the contribution of the underlying repair module.

\noindent\textbf{The Results under Different Missing Rates.}
Table~\ref{tab:missing_rate_acc} reports dataset-specific accuracy across missing rates $MR \in \{0.0, 0.1, \dots, 0.7\}$, using Acc-2 for CMU-MOSI and Acc-4 for IEMOCAP.
Adding SIEVE consistently improves all three repair backbones at every missing rate on both datasets. Two patterns emerge from these results. First, the improvements are not confined to incomplete inputs: SIEVE also helps in the full-modality setting ($MR = 0.0$). This indicates that selective routing is beneficial even when all modalities are observed, since the direct branch can avoid the noise introduced by an unnecessary repair step. Second, the relative gains tend to grow with $MR$, and the effect is more pronounced for weaker backbones. This suggests that SIEVE provides larger benefits when repair signals become less reliable, where deciding \emph{whether} to repair matters more than \emph{how} to repair. 

\begin{table*}[t]
\centering
\small
\setlength{\tabcolsep}{5.2pt}
\renewcommand{\arraystretch}{1.22}
\caption{Accuracy under different missing rates on CMU-MOSI and IEMOCAP.}
\label{tab:missing_rate_acc}
\begin{adjustbox}{max width=\linewidth}
\begin{tabular}{c|l|*{9}{c}}
\hline
Dataset & Method 
& 0.0 & 0.1 & 0.2 & 0.3 & 0.4 & 0.5 & 0.6 & 0.7 & Avg. \\
\hline

\multirow{6}{*}{CMU-MOSI}
& GCNet 
& 0.8445 & 0.8262 & 0.8034 & 0.8171 & 0.7546 & 0.7530 & 0.7119 & 0.7027 & 0.7767 \\
& \cellcolor{ourgray}\quad + SIEVE 
& \cellcolor{ourgray}0.8567 & \cellcolor{ourgray}0.8430 & \cellcolor{ourgray}0.8186 & \cellcolor{ourgray}\textbf{0.8201} & \cellcolor{ourgray}0.7683 & \cellcolor{ourgray}\textbf{0.7744} & \cellcolor{ourgray}\textbf{0.7393} & \cellcolor{ourgray}\textbf{0.7165} & \cellcolor{ourgray}0.7921 \\
\cline{2-11}
& MoMKE 
& 0.8708 & 0.8490 & 0.8340 & 0.7935 & 0.7642 & 0.7410 & 0.7202 & 0.6705 & 0.7804 \\
& \cellcolor{ourgray}\quad + SIEVE 
& \cellcolor{ourgray}\textbf{0.8872} & \cellcolor{ourgray}\textbf{0.8643} & \cellcolor{ourgray}\textbf{0.8460} & \cellcolor{ourgray}0.8140 & \cellcolor{ourgray}\textbf{0.7774} & \cellcolor{ourgray}0.7713 & \cellcolor{ourgray}0.7317 & \cellcolor{ourgray}0.6982 & \cellcolor{ourgray}\textbf{0.7988} \\
\cline{2-11}
& CorrKD 
& 0.8491 & 0.7973 & 0.7546 & 0.7134 & 0.6723 & 0.6250 & 0.6052 & 0.5854 & 0.7003 \\
& \cellcolor{ourgray}\quad + SIEVE 
& \cellcolor{ourgray}0.8567 & \cellcolor{ourgray}0.8018 & \cellcolor{ourgray}0.7500 & \cellcolor{ourgray}0.7226 & \cellcolor{ourgray}0.7043 & \cellcolor{ourgray}0.6799 & \cellcolor{ourgray}0.6372 & \cellcolor{ourgray}0.6174 & \cellcolor{ourgray}0.7212 \\
\hline

\multirow{6}{*}{IEMOCAP}
& GCNet 
& 0.7618 & 0.7524 & 0.7426 & 0.7319 & 0.7198 & 0.7076 & 0.6928 & 0.6816 & 0.7238 \\
& \cellcolor{ourgray}\quad + SIEVE 
& \cellcolor{ourgray}0.7737 & \cellcolor{ourgray}0.7651 & \cellcolor{ourgray}\textbf{0.7540} & \cellcolor{ourgray}\textbf{0.7452} & \cellcolor{ourgray}\textbf{0.7310} & \cellcolor{ourgray}\textbf{0.7204} & \cellcolor{ourgray}\textbf{0.7031} & \cellcolor{ourgray}\textbf{0.6942} & \cellcolor{ourgray}\textbf{0.7358} \\
\cline{2-11}
& MoMKE 
& 0.7760 & 0.7512 & 0.7362 & 0.7147 & 0.6954 & 0.6840 & 0.6651 & 0.6437 & 0.7083 \\
& \cellcolor{ourgray}\quad + SIEVE 
& \cellcolor{ourgray}\textbf{0.7925} & \cellcolor{ourgray}\textbf{0.7682} & \cellcolor{ourgray}0.7478 & \cellcolor{ourgray}0.7252 & \cellcolor{ourgray}0.7108 & \cellcolor{ourgray}0.6914 & \cellcolor{ourgray}0.6805 & \cellcolor{ourgray}0.6555 & \cellcolor{ourgray}0.7215 \\
\cline{2-11}
& CorrKD 
& 0.7075 & 0.6778 & 0.6566 & 0.6278 & 0.6120 & 0.5894 & 0.5742 & 0.5575 & 0.6254 \\
& \cellcolor{ourgray}\quad + SIEVE 
& \cellcolor{ourgray}0.7270 & \cellcolor{ourgray}0.6990 & \cellcolor{ourgray}0.6808 & \cellcolor{ourgray}0.6576 & \cellcolor{ourgray}0.6318 & \cellcolor{ourgray}0.6131 & \cellcolor{ourgray}0.5890 & \cellcolor{ourgray}0.5756 & \cellcolor{ourgray}0.6467 \\
\hline
\end{tabular}
\end{adjustbox}
\end{table*}

\noindent\textbf{Ablation Study.}
We conduct ablations on CMU-MOSI with GCNet as the repair backbone, and report Acc-2 under three representative missing rates in Table~\ref{tab:ablation}. Removing the valve calibration loss $\mathcal{L}_{\mathrm{valve}}$ degrades performance at all missing rates, confirming that the Beta likelihood over $s_i^\star$ is essential for learning a useful valve. Removing the evidence regularizer $\mathcal{L}_{\mathrm{ev}}$ causes the largest drop, indicating that penalizing wrong-direction evidence stabilizes uncertainty estimation. Removing the repair-module loss $\mathcal{L}_{R}$ also hurts performance, showing that selective routing still relies on a well-trained repair branch. On the routing side, replacing $\theta(u_i)$ with a fixed threshold $\theta = 0.5$ removes uncertainty adaptivity and consistently lowers accuracy, while fixing $\pi_{m} = 0$ (which reduces Eq.~\eqref{eq:threshold} to $\theta(u_i) = (1 + u_i)/2$ and imposes a uniformly conservative prior across missing patterns) also degrades performance. Together, these results validate the three pillars of SIEVE: empirical sufficiency supervision, evidential regularization, and pattern-aware uncertainty-adaptive routing.
\vspace{-0.2cm}
\begin{table}[H]
\centering
\small
\caption{Ablation study under different missing rates on the CMU-MOSI dataset in terms of Acc-2.}
\vspace{-0.2cm}
\label{tab:ablation}
\begin{tabular}{lccc}
\toprule
\textbf{Variant} & \textbf{$MR=0.3$} & \textbf{$MR=0.5$} & \textbf{$MR=0.7$} \\
\midrule
Full & \textbf{0.8201} & \textbf{0.7744} & \textbf{0.7165} \\
\midrule
w/o $\mathcal{L}_{\mathrm{valve}}$ & 0.7530 & 0.6707 & 0.6204 \\
w/o $\mathcal{L}_{\mathrm{ev}}$ & 0.7378 & 0.6128 & 0.6250 \\
w/o $\mathcal{L}_{R}$ & 0.7302 & 0.6723 & 0.6174 \\
\midrule
fixed $\theta=0.5$ & 0.7409 & 0.6738 & 0.6128 \\
fixed $\pi_m=0$ & 0.7500 & 0.6357 & 0.6189 \\
\bottomrule
\end{tabular}
\end{table}
\vspace{-0.2cm}

\noindent\textbf{Anatomy of SIEVE's Routing Behavior.}
\label{sec:routing_analysis}
We investigate two questions: (i) does the evidential valve learn a meaningful sufficiency signal, and (ii) how much of the achievable accuracy does it recover? We study both on CMU-MOSI under missing rates $MR \in \{0.3, 0.5, 0.7\}$. For each test utterance, we record its missing pattern $m_i$ and predicted sufficiency $\bar{s}_i = \alpha_i / (\alpha_i + \beta_i)$. As an upper reference, we compute the \emph{Achievable Optimum}, which selects, for each sample, the branch with the smaller per-sample loss; this denotes the highest accuracy reachable by any binary router restricted to the two branches.

\begin{figure}[t]
\centering
\begin{subfigure}{0.32\linewidth}
  \includegraphics[width=\linewidth]{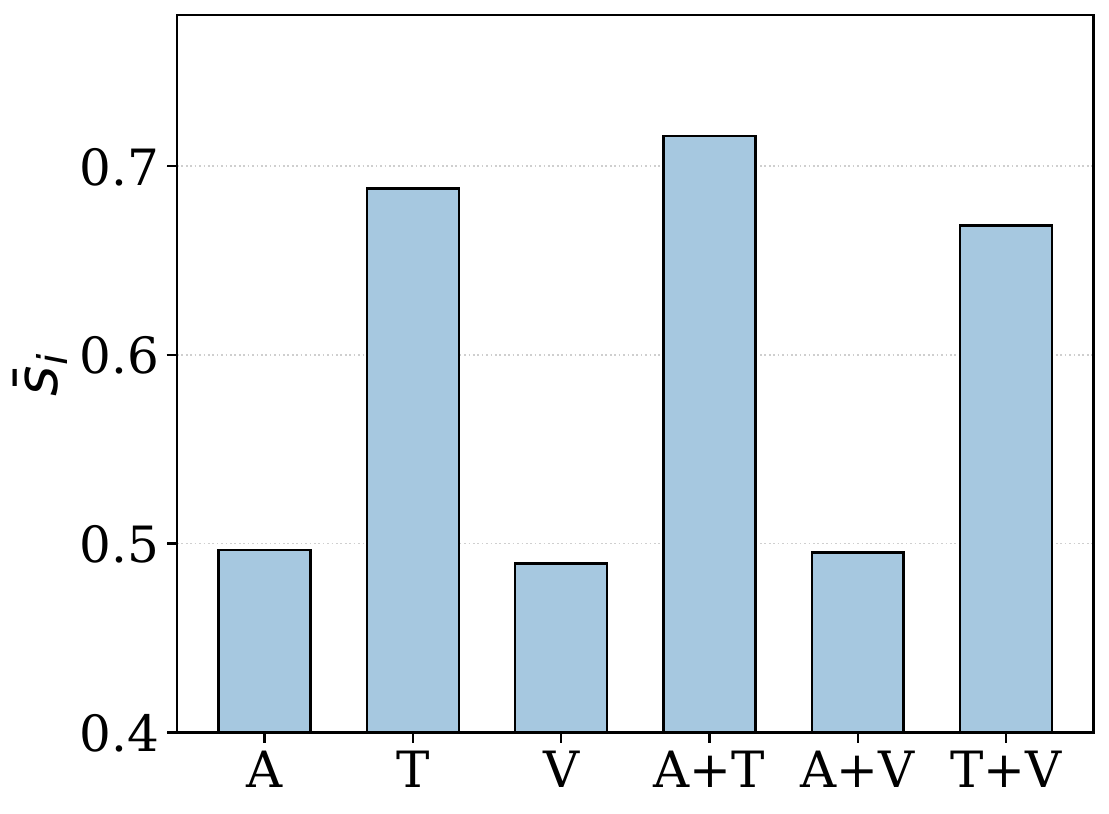}
  \caption{$\bar{s}_i$ at $MR=0.3$}
  \label{fig:sbar_03}
\end{subfigure}\hfill
\begin{subfigure}{0.32\linewidth}
  \includegraphics[width=\linewidth]{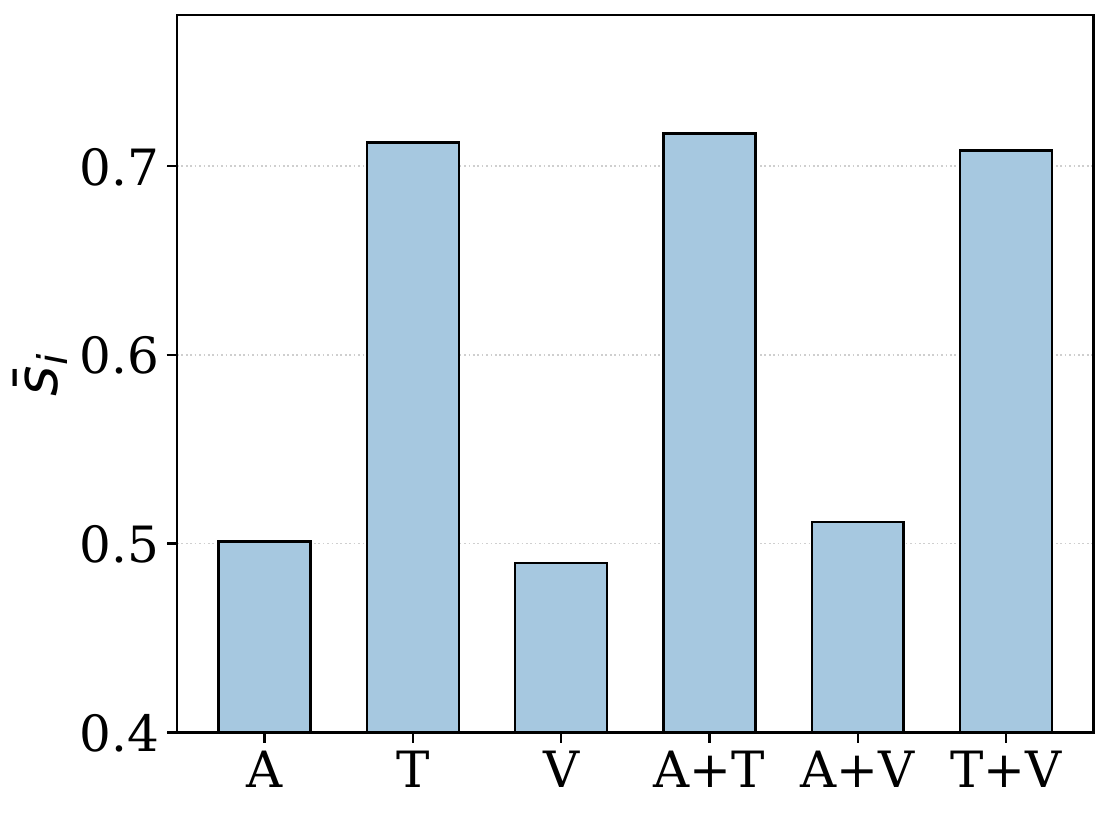}
  \caption{$\bar{s}_i$ at $MR=0.5$}
  \label{fig:sbar_05}
\end{subfigure}\hfill
\begin{subfigure}{0.32\linewidth}
  \includegraphics[width=\linewidth]{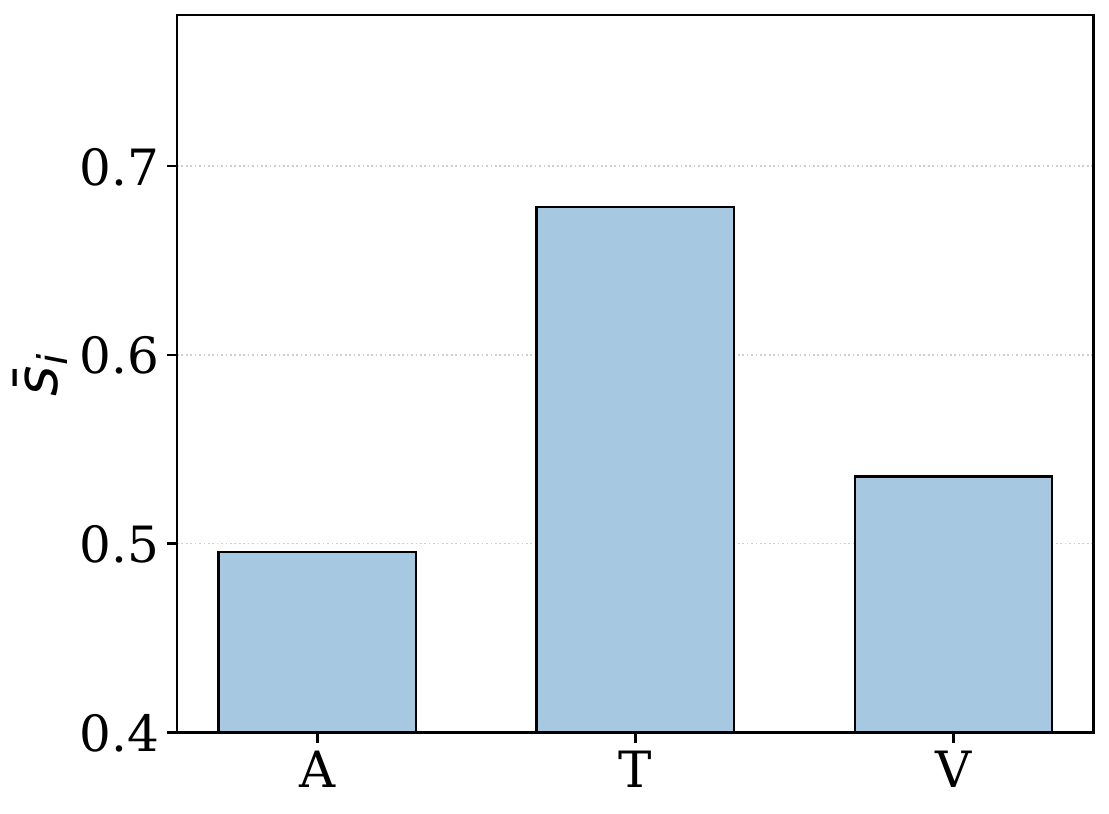}
  \caption{$\bar{s}_i$ at $MR=0.7$}
  \label{fig:sbar_07}
\end{subfigure}

\vspace{0.4em}

\begin{subfigure}{0.32\linewidth}
  \includegraphics[width=\linewidth]{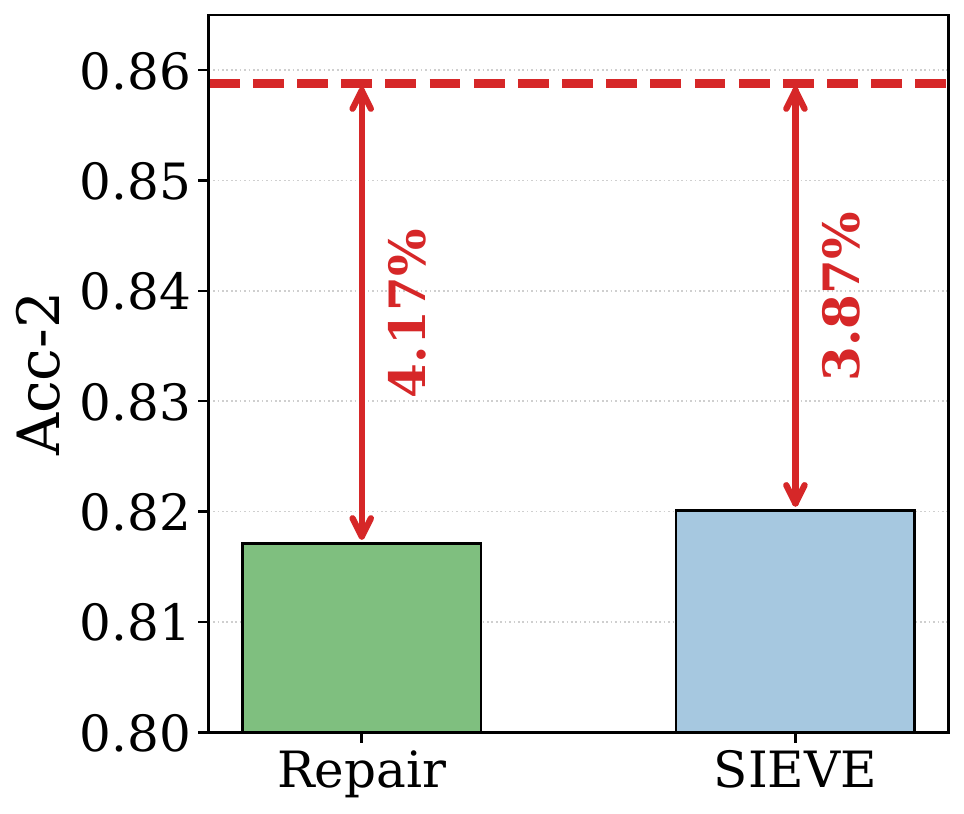}
  \caption{Gap to optimum at $MR=0.3$}
  \label{fig:opt_03}
\end{subfigure}\hfill
\begin{subfigure}{0.32\linewidth}
  \includegraphics[width=\linewidth]{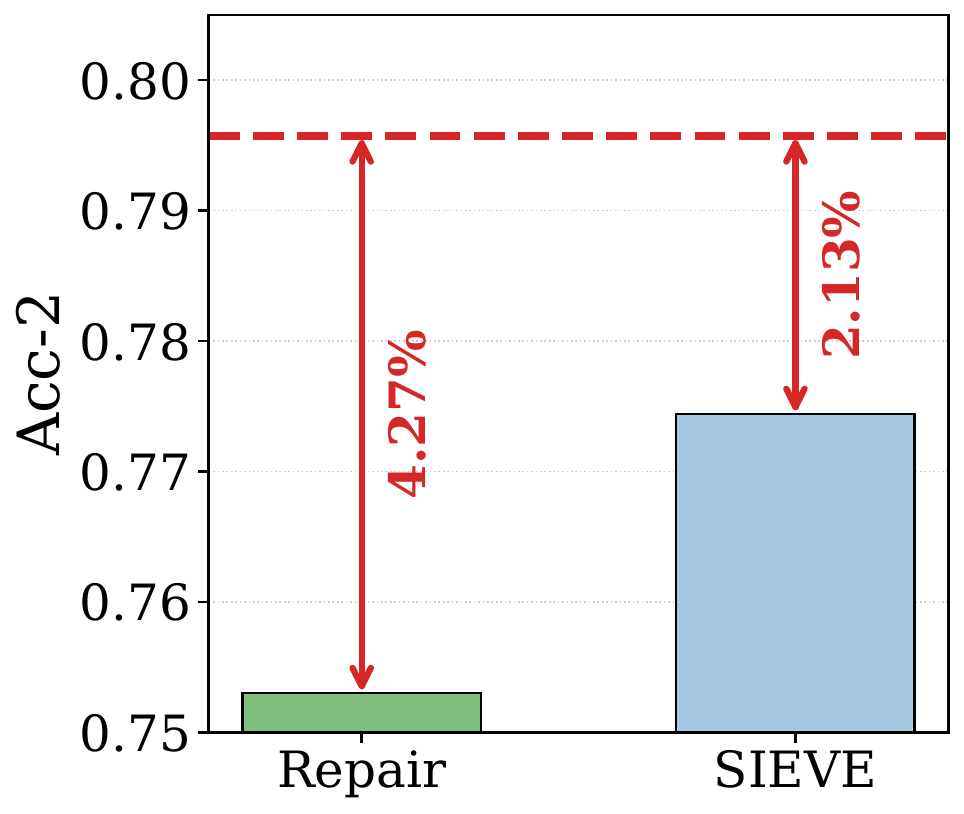}
  \caption{Gap to optimum at $MR=0.5$}
  \label{fig:opt_05}
\end{subfigure}\hfill
\begin{subfigure}{0.32\linewidth}
  \includegraphics[width=\linewidth]{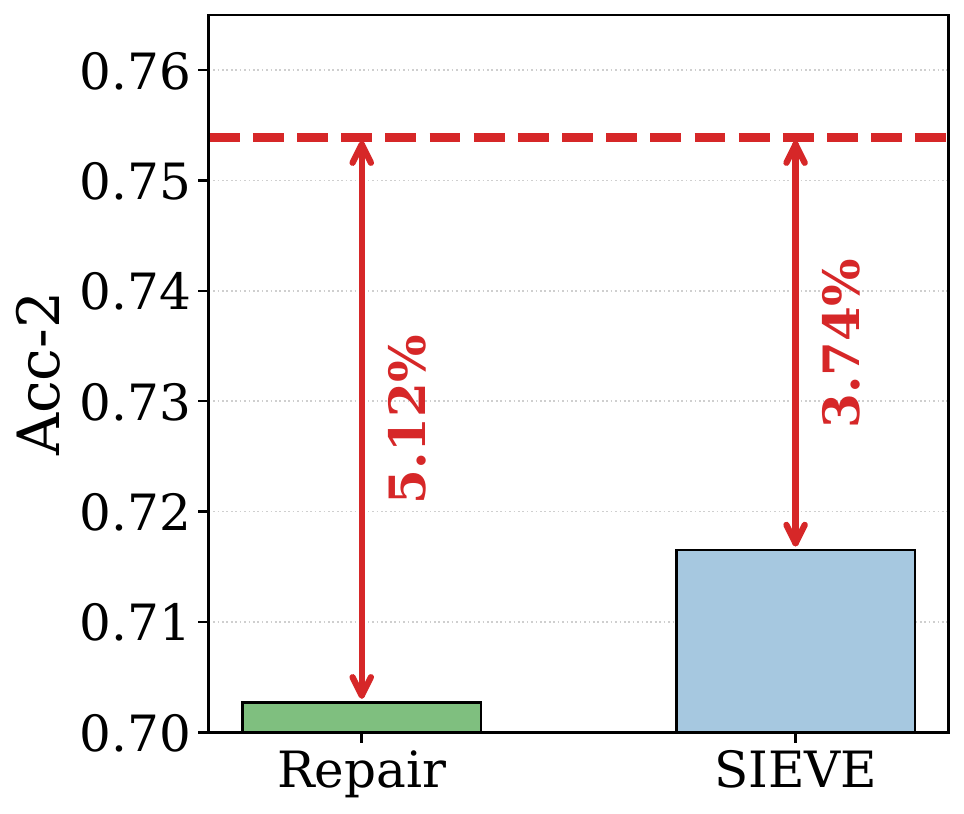}
  \caption{Gap to optimum at $MR=0.7$}
  \label{fig:opt_07}
\end{subfigure}
\vspace{-0.2cm}
\caption{Routing analysis on CMU-MOSI under three missing rates.
\textbf{Top row}: mean per-pattern sufficiency $\bar{s}_i$ produced by SIEVE.
\textbf{Bottom row}: Acc-2 of Repair-only and SIEVE; the red dashed line marks the per-sample Achievable Optimum.
}
\label{fig:routing}
\end{figure}

Figure~\ref{fig:routing} reveals three findings. First, $\bar{s}_i$ is consistently higher for patterns containing the language modality, reflecting the text-dominant nature of sentiment evidence on CMU-MOSI and matching the oracle distribution in Figure~\ref{fig:oracle}. Second, this ranking remains stable across missing rates, suggesting that the valve learns a transferable notion of sufficiency; meanwhile, the sufficiency of language-free patterns rises with $MR$, showing that the valve adaptively raises its tolerance for direct prediction when repair becomes less reliable. Third, SIEVE consistently outperforms the Repair-only baseline and approaches the Achievable Optimum across all three regimes, closing most of the gap to the per-sample upper bound.

\noindent\textbf{Branch Representations Analysis.}
\label{sec:tsne_analysis}
A natural concern is whether the direct and repair branches collapse into redundant representations, in which case per-sample routing would degenerate into a learned ensemble weight. We therefore ask: do the two branches occupy distinct regions of the representation space, and does this separation persist as missingness intensifies? For each test utterance on CMU-MOSI, we extract the penultimate-layer feature of the routed branch and project it into two dimensions via t-SNE, coloring each point by its hard routing decision $g_i^{\mathrm{hard}}$. The analysis is repeated at $MR \in \{0.3, 0.5, 0.7\}$.

\begin{figure}[t]
\centering
\begin{subfigure}{0.32\linewidth}
  \includegraphics[width=\linewidth]{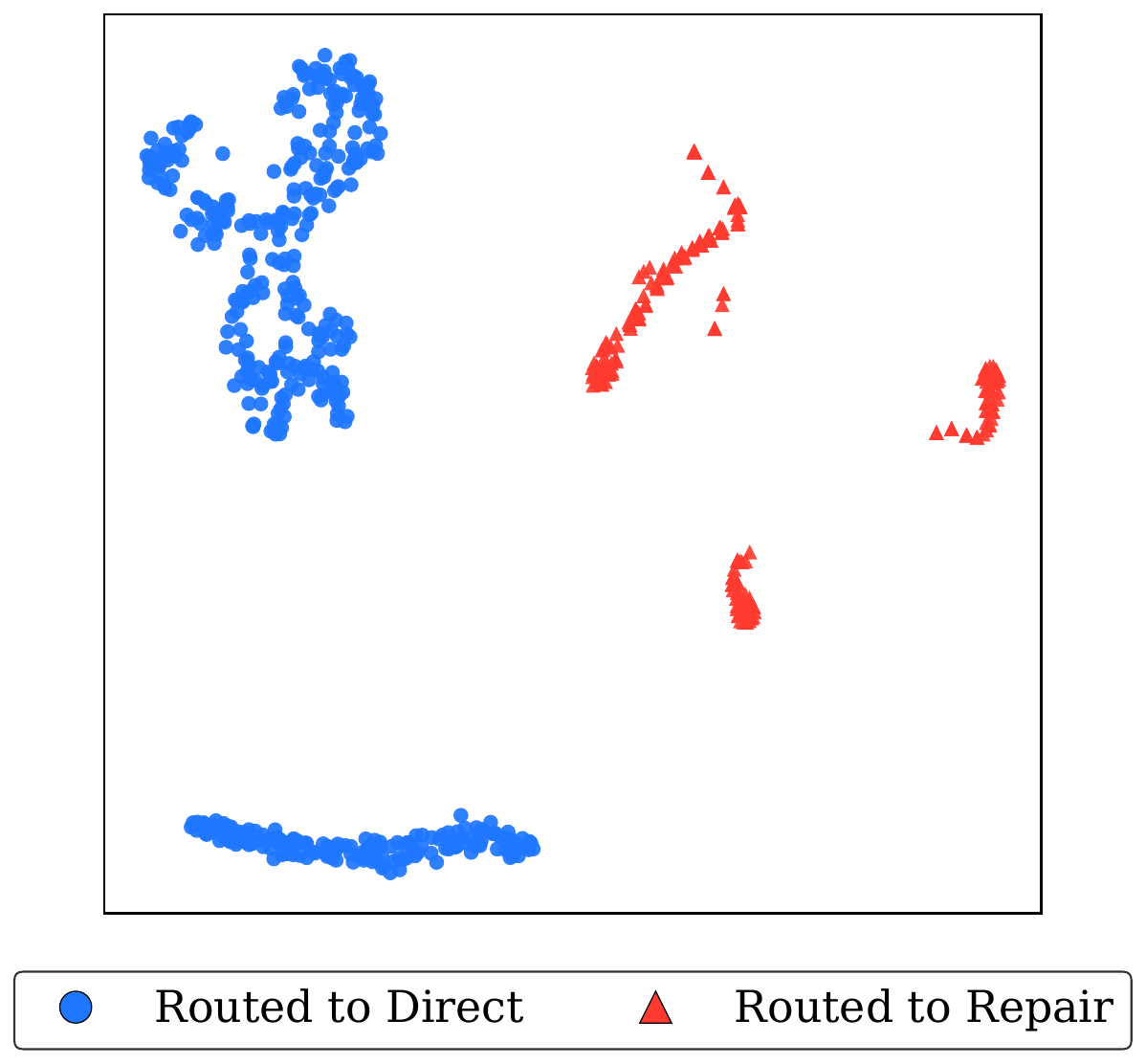}
  \caption{$MR=0.3$}
  \label{fig:tsne_03}
\end{subfigure}\hfill
\begin{subfigure}{0.32\linewidth}
  \includegraphics[width=\linewidth]{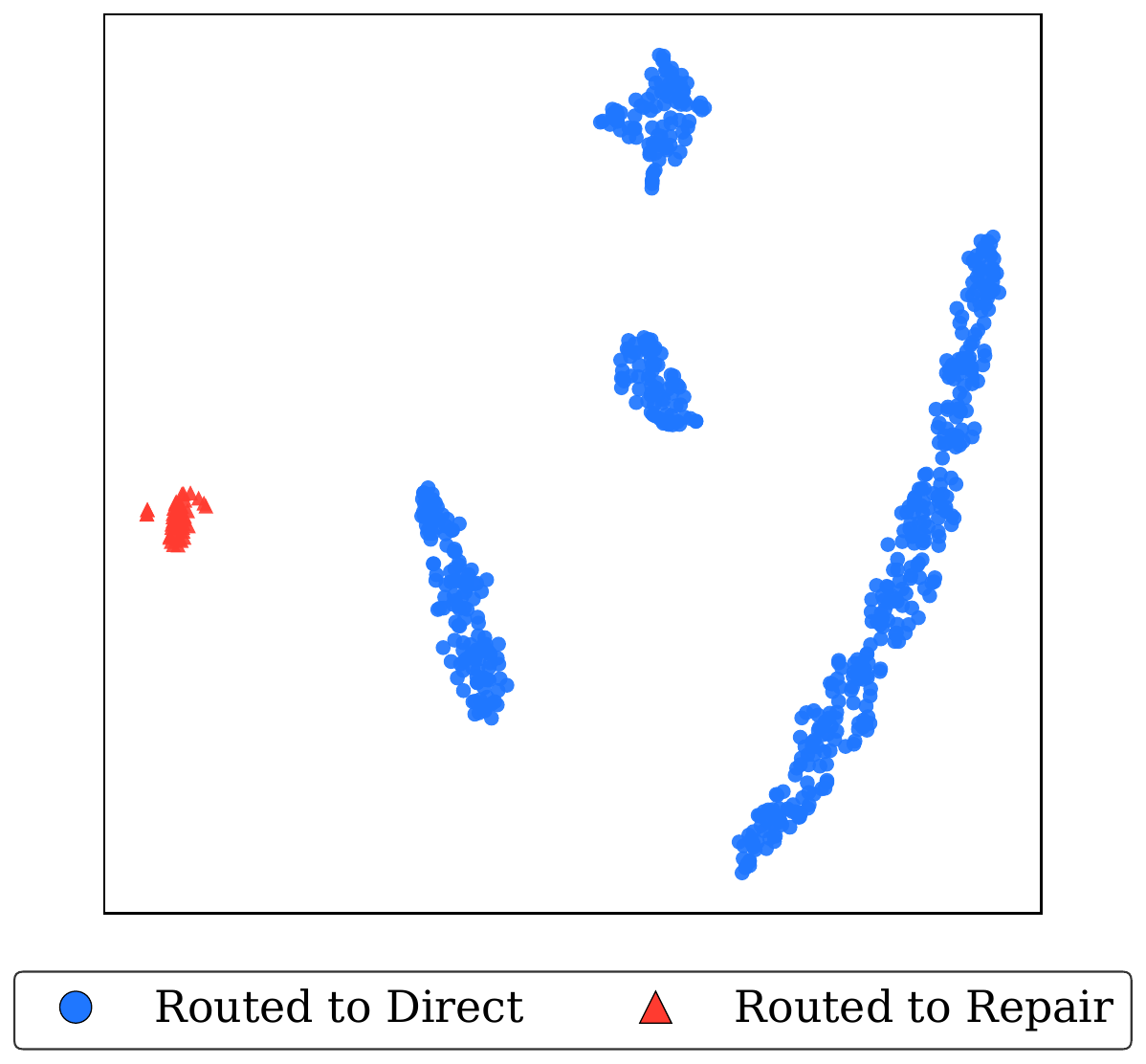}
  \caption{$MR=0.5$}
  \label{fig:tsne_05}
\end{subfigure}\hfill
\begin{subfigure}{0.32\linewidth}
  \includegraphics[width=\linewidth]{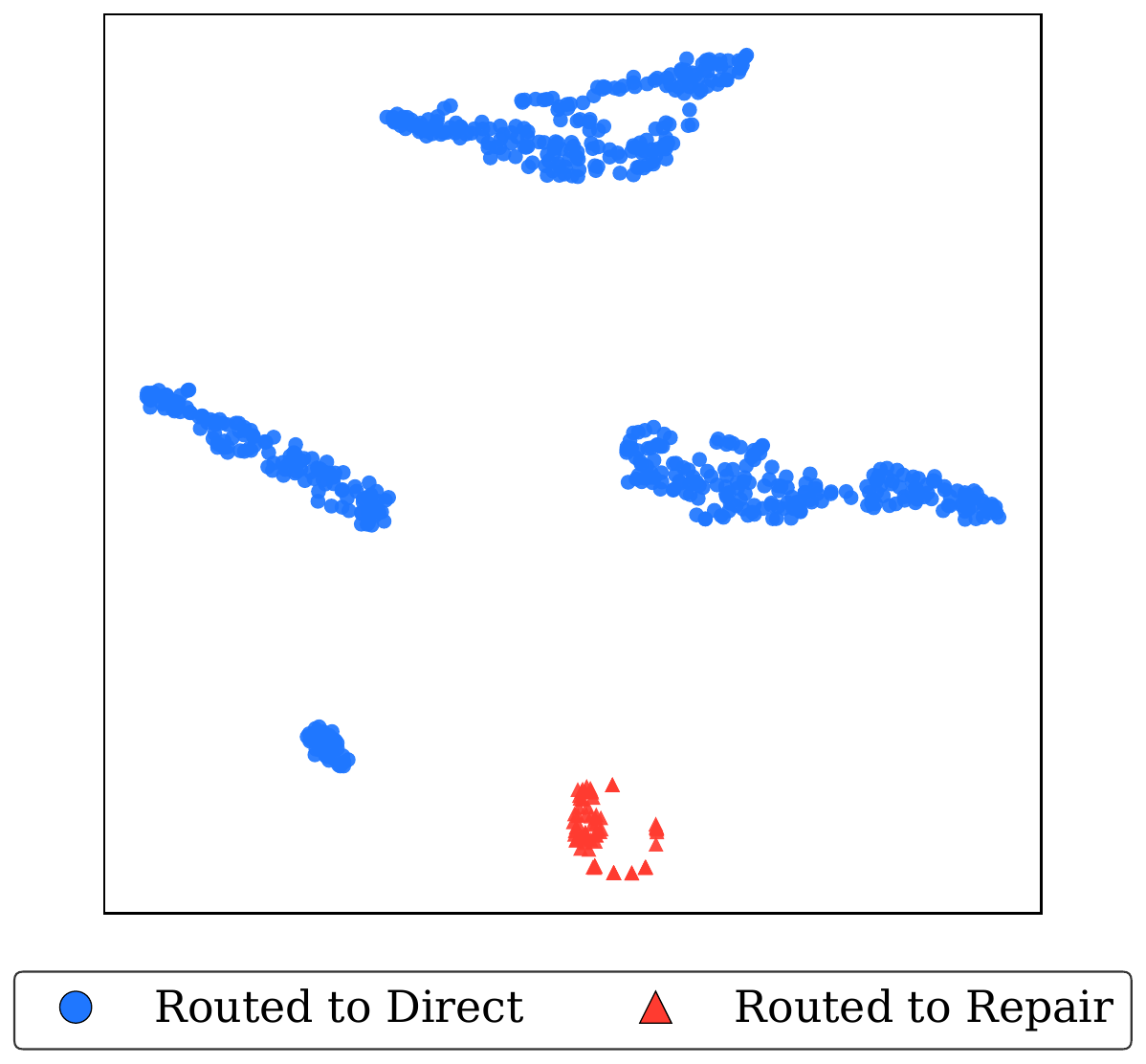}
  \caption{$MR=0.7$}
  \label{fig:tsne_07}
\end{subfigure}
\vspace{-0.2cm}
\caption{t-SNE projection of per-sample routed-branch representations on CMU-MOSI under three missing rates, colored by routing decision.}
\label{fig:tsne}
\end{figure}

Figure~\ref{fig:tsne} shows a clear separation across all three regimes: Direct-routed and Repair-routed samples occupy disjoint regions of the embedding space, indicating that the two pathways encode structurally complementary information rather than duplicating each other. Notably, the Repair-routed cluster shrinks in relative size as $MR$ grows—an initially counterintuitive trend that aligns with our earlier observation: when repair becomes less reliable under heavy missingness, the valve adaptively favors the direct pathway. Together, these results show that SIEVE operates over representationally distinct branches rather than selecting between near-duplicate predictors.

\section{Conclusion}
We challenged the repair-first assumption that underlies most missing-modality MSA research, and showed through a per-sample oracle analysis that the utility of a modality is sample-dependent: repair is not universally beneficial, and is unnecessary for a substantial fraction of samples. Motivated by this observation, we proposed SIEVE, a plug-and-play framework that learns, for each input, whether the observed modalities are already sufficient or whether repair should be invoked. SIEVE turns sufficiency into an internal-supervised quantity through a dual-branch loss gap and routes samples through an evidential gate that jointly accounts for sufficiency and its uncertainty. Across two benchmarks, diverse repair backbones, and a wide range of missing rates, SIEVE consistently improves the underlying repair models and approaches the per-sample achievable optimum, without complete-modality supervision. More broadly, our results establish \emph{whether to repair} as a first-class question alongside \emph{how to repair}, and suggest that selective intervention is a useful design principle for robust multimodal systems beyond sentiment analysis.

\section{Limitations}
Our empirical study is restricted to multimodal sentiment analysis on CMU-MOSI and IEMOCAP. Extending SIEVE to broader incomplete-modality settings, including visual question answering, multimodal action recognition, and medical multimodal diagnosis, is a promising direction for future work.

In addition, SIEVE evaluates both branches during training to compute the per-sample loss gap. The overhead is modest in practice, since the direct branch is lightweight relative to the repair module, but it is nonzero. Future work will explore amortized training schemes and inference-time branch pruning to further reduce this cost.

\bibliography{custom}

\appendix
\section{Derivation of the Adaptive Threshold}
\label{app:threshold}
This appendix shows that the dynamic threshold $\theta(u) = (1 + (1 - 2\pi_m)u) / 2$ in Eq.~\eqref{eq:threshold} arises as the exact decision boundary induced by Subjective Logic projection under a missing-pattern-conditioned base rate. We drop the sample index $i$ throughout.

\paragraph{Beta-opinion correspondence.}
The Beta parameters $\alpha, \beta > 1$ map to a Subjective Logic opinion $(b, d, u)$ with $b + d + u = 1$~\citep{josang2016subjective}:
\begin{equation}
b = \frac{\alpha - 1}{S}, \ d = \frac{\beta - 1}{S}, \quad u = \frac{2}{S}, \quad S = \alpha + \beta.
\end{equation}
Using the Beta mean $\bar{s} = \alpha / S$, this rewrites as
\begin{equation}
b = \bar{s} - \tfrac{u}{2}, \ d = (1 - \bar{s}) - \tfrac{u}{2}.
\label{eq:bd-sbar}
\end{equation}

\paragraph{Missing-pattern-conditioned projection.}
A Subjective Logic opinion is projected to a decision probability via a base rate $\pi \in [0, 1]$:
\begin{equation}
P_s = b + \pi\, u, \ P_{\neg s} = d + (1 - \pi)\, u.
\end{equation}
A fixed base rate treats every missing pattern identically, which is overly pessimistic for patterns where the observed modalities are usually sufficient, and overly conservative for patterns where they are rarely sufficient. We therefore condition the base rate on the missing pattern $m$, defining $\pi_m$ as the empirical probability that direct prediction is at least as accurate as repair under pattern $m$. Operationally, $\pi_m$ is estimated online from the empirical sufficiency target $s^\star$ defined in Eq.(9) via an exponential moving average (Eq.(15)), requiring no additional external supervision. The projection becomes
\begin{equation}
P_s = b + \pi_m\, u, \ P_{\neg s} = d + (1 - \pi_m)\, u.
\label{eq:pi-projection}
\end{equation}

\paragraph{Threshold.}
Routing to the direct branch requires $P_s > P_{\neg s}$. Substituting Eq.~\eqref{eq:bd-sbar} and Eq.~\eqref{eq:pi-projection},
\begin{equation}
\begin{split}
&\bar{s} - \tfrac{u}{2} + \pi_m\, u > (1 - \bar{s}) - \tfrac{u}{2} + (1 - \pi_m)\, u \\
&\qquad \Longleftrightarrow\; \bar{s} > \frac{1 + (1 - 2\pi_m)\, u}{2},
\end{split}
\end{equation}
which is exactly $\bar{s} > \theta(u)$. The threshold decomposes as $\theta(u) = \tfrac{1}{2} + (\tfrac{1}{2} - \pi_m)\,u$, where the constant $\tfrac{1}{2}$ is the decision boundary under full confidence ($u \to 0$) and $(\tfrac{1}{2} - \pi_m)\,u$ is an uncertainty-scaled correction whose direction is set by the missing-pattern base rate. When $\pi_m > \tfrac{1}{2}$, the correction is negative and the threshold drops below $\tfrac{1}{2}$, allowing the valve to more readily skip repair on patterns that are usually sufficient; when $\pi_m < \tfrac{1}{2}$, the threshold rises above $\tfrac{1}{2}$, demanding stronger sufficiency evidence before skipping repair. The magnitude of this shift scales with $u$: under low uncertainty, the threshold stays close to $\tfrac{1}{2}$, while under high uncertainty it approaches $1 - \pi_m$, deferring the routing decision to the missing-pattern prior rather than to a fixed default.
\section{Extended Related Work: MSA with Missing Modalities}
\label{app:related-work}

Multimodal sentiment analysis (MSA) integrates language, acoustic, and visual signals to infer human affect. Since real-world inputs are often incomplete, recent studies have explored robust MSA under missing modalities. From a repair perspective, these methods can be broadly grouped into explicit and implicit modality repair.

\paragraph{Explicit modality repair.}
These methods reconstruct, imagine, or approximate absent modalities or their latent representations from the observed ones before prediction. Early studies learn cross-modal translation or imagination mechanisms~\cite{pham2019found,zhao2021missing}, while later work performs graph-based completion for conversational data~\cite{lian2023gcnet}. Recent methods further improve recovery with more expressive generators, including prompt-based generation~\cite{guo2024multimodal}, class-specific normalizing flows~\cite{wang2023distribution}, and conditional diffusion~\cite{wang2023incomplete}.

\paragraph{Implicit modality repair.}
Rather than directly filling the missing slots, these methods compensate for the resulting semantic deficiency at the representation or decision level. One line of work transfers full-modality knowledge to incomplete-modality models via distillation, including margin-aware decision distillation~\cite{wei2023mmanet}, self-distillation across heterogeneous missing cases~\cite{li2024unified}, hierarchical representation alignment~\cite{li2024toward}, multi-level correlation distillation~\cite{li2024correlation}, and modality-aware feature distillation~\cite{zhuang2025cmad}. The other line performs intrinsic evidence compensation without a full-modality teacher, by refining a language-dominant signal~\cite{zhang2024towards}, retrieving cross-sample cues~\cite{zhuang2025hyper}, combining modality-expert knowledge~\cite{xu2024leveraging}, or denoising modality-specific and shared representations~\cite{zhuang2026tmdc}.

\paragraph{Position of SIEVE.}
Overall, existing methods focus on \emph{how} to repair missing modalities, applying their compensation mechanism uniformly to every incomplete sample. In contrast, SIEVE asks \emph{whether} repair is needed for each sample, and selectively activates the underlying repair module only when the observed modalities are insufficient. A concurrent line of work~\citep{chen2026evaluation,bi2025two} begins to question the universality of repair, but differs from SIEVE in two essential aspects. First, their sufficiency signal depends on external supervision: ProMMA~\cite{chen2026evaluation} derives pseudo-labels from a complete-modality teacher and a hand-tuned similarity threshold, while DREAM~\cite{bi2025two} relies on ground-truth labels and requires enumerating all $2^{|\mathcal{M}|}$ modality subsets to compute its Shapley-based contribution score. SIEVE instead grounds sufficiency in the dual-branch loss gap, which is fully self-contained and requires only a single forward pass per branch. Second, their selection mechanism is tightly entangled with a specific repair design: ProMMA's evaluator operates on its own prompt-based generator and cannot be transplanted without rebuilding the prompt modules, while DREAM's PSS-driven enhancement and soft-masking fusion are co-designed and inseparable from its cross-modal attention backbone. SIEVE imposes no constraint on $\mathcal{R}_\omega$ and acts as a unified decision layer that wraps any repair module as a black box, which we validate by attaching the same SIEVE design to three repair paradigms spanning explicit reconstruction (GCNet), distillation-based compensation (CorrKD), and intrinsic evidence compensation (MoMKE).

\section{Hyperparameter Settings}
\label{app:hyperparameters}

For all backbone models, we follow the experimental settings described in
Section~\ref{sec:exp_setup}. The feature extractors, missing-modality protocol,
dataset splits, and evaluation metrics are kept identical between each backbone
and its SIEVE-enhanced counterpart. We do not modify the internal repair
architecture of the backbone models when adding SIEVE. Instead, SIEVE only
introduces a lightweight evidential valve on top of the existing prediction
pipeline, together with the valve-related losses described in the method
section.

Table~\ref{tab:sieve-hyperparameters} reports the SIEVE-specific
hyperparameters used in the main incomplete-modality setting ($MR=0.7$). Here,
each method denotes the corresponding repair backbone equipped with SIEVE, while
the original backbone hyperparameters are kept the same as in the main
experiments. The reported hyperparameters include the valve calibration weight
$\lambda_v$, the evidence regularization weight $\lambda_e$, the soft
sufficiency-label temperature $\kappa$, the final Gumbel-Sigmoid temperature
$\tau_{\min}$, and the hidden dimension of the valve MLP. The Gumbel-Sigmoid
temperature is linearly annealed from $\tau_0=1.0$ to $\tau_{\min}$ during
training.

The values are chosen from a compact and shared range across datasets and
backbones. In particular, $\lambda_v$ controls how strongly the empirical
direct--repair loss gap supervises the valve; $\lambda_e$ controls the strength
of evidential regularization; $\kappa$ determines how sharply the loss gap is
converted into the soft sufficiency target; and $\tau_{\min}$ controls the
degree to which the learned routing approaches a near-binary decision late in
training. We keep the gate hidden dimension small relative to the backbone
encoders, so the additional parameter cost of SIEVE remains negligible.

\begin{table}[t]
\centering
\small
\caption{SIEVE-specific hyperparameters used under $MR=0.7$.}
\label{tab:sieve-hyperparameters}
\setlength{\tabcolsep}{3.2pt}
\renewcommand{\arraystretch}{2.2}
\begin{tabular}{llccccc}
\toprule
\textbf{Data} & \textbf{Method} 
& $\lambda_v$ 
& $\lambda_e$ 
& $\kappa$ 
& $\tau_{\min}$ 
& \textbf{Valve Dimension} \\
\midrule
MOSI & GCNet  & 0.05 & 0.10 & 0.50 & 0.10 & 128 \\
MOSI & MoMKE  & 0.30 & 0.01 & 0.30 & 0.10 & 128 \\
MOSI & CorrKD & 0.30 & 0.01 & 0.30 & 0.10 & 128 \\
IEMO & GCNet  & 0.10 & 0.05 & 0.30 & 0.20 & 128 \\
IEMO & MoMKE  & 0.15 & 0.06 & 0.12 & 0.30 & 96  \\
IEMO & CorrKD & 0.12 & 0.12 & 0.08 & 0.45 & 96  \\
\bottomrule
\end{tabular}
\end{table}

\section{Full Results under Different Missing Rates}
\label{app:full_results}

To complement the accuracy-only comparison in Table~\ref{tab:missing_rate_acc},
we report the full evaluation results under all missing rates in this appendix.
For CMU-MOSI, we include Acc-2, weighted F1, MAE, and Pearson correlation.
For IEMOCAP, which is evaluated as a four-class classification task, we report
Acc-4 and weighted F1. These results provide a more complete view of how SIEVE
affects both classification-oriented and regression-oriented evaluation
criteria.

\begin{table*}[p]
\centering
\small
\setlength{\tabcolsep}{2.2pt}
\renewcommand{\arraystretch}{2.2}
\caption{Full results under different missing rates.}
\label{tab:app_full_results}

\noindent\textbf{(a) CMU-MOSI results. Each cell reports Acc-2/F1.}
\vspace{0.3em}

\resizebox{\textwidth}{!}{%
\begin{tabular}{lccccccccc}
\toprule
\textbf{Method}
& \textbf{0.0} & \textbf{0.1} & \textbf{0.2} & \textbf{0.3}
& \textbf{0.4} & \textbf{0.5} & \textbf{0.6} & \textbf{0.7}
& \textbf{Avg.} \\
\midrule
GCNet
& 0.8445/0.8442 & 0.8262/0.8244 & 0.8034/0.8033 & 0.8171/0.8164
& 0.7546/0.7560 & 0.7530/0.7532 & 0.7119/0.7127 & 0.7027/0.6998
& 0.7767/0.7762 \\
\rowcolor{ourgray}
\quad + SIEVE
& 0.8567/0.8569 & 0.8430/0.8425 & 0.8186/0.8188 & 0.8201/0.8164
& 0.7683/0.7684 & 0.7744/0.7758 & 0.7393/0.7396 & 0.7165/0.7100
& 0.7921/0.7911 \\
\midrule
CorrKD
& 0.8491/0.8491 & 0.7973/0.7982 & 0.7546/0.7551 & 0.7134/0.7133
& 0.6723/0.6689 & 0.6250/0.6121 & 0.6052/0.5904 & 0.5854/0.5642
& 0.7003/0.6939 \\
\rowcolor{ourgray}
\quad + SIEVE
& 0.8567/0.8568 & 0.8018/0.8030 & 0.7500/0.7515 & 0.7226/0.7218
& 0.7043/0.7015 & 0.6799/0.6736 & 0.6372/0.6238 & 0.6174/0.5993
& 0.7212/0.7164 \\
\midrule
MoMKE
& 0.8708/0.8705 & 0.8490/0.8492 & 0.8340/0.8346 & 0.7935/0.7942
& 0.7642/0.7653 & 0.7410/0.7424 & 0.7202/0.7218 & 0.6705/0.6718
& 0.7804/0.7812 \\
\rowcolor{ourgray}
\quad + SIEVE
& 0.8872/0.8872 & 0.8643/0.8647 & 0.8460/0.8460 & 0.8140/0.8146
& 0.7774/0.7788 & 0.7713/0.7727 & 0.7317/0.7333 & 0.6982/0.6999
& 0.7988/0.7996 \\
\bottomrule
\end{tabular}%
}

\vspace{9.2em}

\noindent\textbf{(b) CMU-MOSI results. Each cell reports MAE/Corr. Lower MAE is better, while higher Corr is better.}
\vspace{0.3em}

\resizebox{\textwidth}{!}{%
\begin{tabular}{lccccccccc}
\toprule
\textbf{Method}
& \textbf{0.0} & \textbf{0.1} & \textbf{0.2} & \textbf{0.3}
& \textbf{0.4} & \textbf{0.5} & \textbf{0.6} & \textbf{0.7}
& \textbf{Avg.} \\
\midrule
GCNet
& 0.8356/0.7698 & 0.8515/0.7341 & 0.9041/0.7185 & 0.9323/0.6960
& 1.0533/0.6094 & 1.0174/0.6364 & 1.1300/0.5644 & 1.2312/0.4910
& 0.9944/0.6524 \\
\rowcolor{ourgray}
\quad + SIEVE
& 0.7852/0.7833 & 0.8085/0.7699 & 0.9199/0.7037 & 0.8996/0.7093
& 1.0072/0.6278 & 1.0553/0.6447 & 1.0800/0.5827 & 1.1734/0.4832
& 0.9661/0.6631 \\
\midrule
CorrKD
& 0.7547/0.7985 & 0.8442/0.7422 & 0.9082/0.7193 & 0.9583/0.6778
& 1.0318/0.6213 & 1.1417/0.5631 & 1.1552/0.5323 & 1.2047/0.4819
& 0.9998/0.6421 \\
\rowcolor{ourgray}
\quad + SIEVE
& 0.7439/0.8105 & 0.7750/0.8047 & 0.9045/0.7090 & 1.0000/0.6436
& 1.0389/0.6150 & 1.0879/0.5819 & 1.1393/0.5414 & 1.1594/0.5321
& 0.9811/0.6548 \\
\midrule
MoMKE
& 0.7828/0.7980 & 0.8372/0.7640 & 0.8905/0.7255 & 0.9750/0.6768
& 1.0398/0.6275 & 1.1020/0.5742 & 1.1480/0.5458 & 1.2235/0.4660
& 0.9998/0.6472 \\
\rowcolor{ourgray}
\quad + SIEVE
& 0.7573/0.8125 & 0.8306/0.7672 & 0.8575/0.7413 & 0.9332/0.6952
& 1.0194/0.6272 & 1.0767/0.5916 & 1.1222/0.5563 & 1.1836/0.4821
& 0.9726/0.6592 \\
\bottomrule
\end{tabular}%
}

\vspace{9.2em}

\noindent\textbf{(c) IEMOCAP results. Each cell reports Acc-4/F1.}
\vspace{0.3em}

\resizebox{\textwidth}{!}{%
\begin{tabular}{lccccccccc}
\toprule
\textbf{Method}
& \textbf{0.0} & \textbf{0.1} & \textbf{0.2} & \textbf{0.3}
& \textbf{0.4} & \textbf{0.5} & \textbf{0.6} & \textbf{0.7}
& \textbf{Avg.} \\
\midrule
GCNet
& 0.7618/0.7642 & 0.7524/0.7549 & 0.7426/0.7453 & 0.7319/0.7341
& 0.7198/0.7224 & 0.7076/0.7101 & 0.6928/0.6955 & 0.6816/0.6840
& 0.7238/0.7263 \\
\rowcolor{ourgray}
\quad + SIEVE
& 0.7737/0.7759 & 0.7651/0.7674 & 0.7540/0.7562 & 0.7452/0.7475
& 0.7310/0.7338 & 0.7204/0.7227 & 0.7031/0.7056 & 0.6942/0.6968
& 0.7358/0.7382 \\
\midrule
CorrKD
& 0.7075/0.7070 & 0.6778/0.6772 & 0.6566/0.6560 & 0.6278/0.6284
& 0.6120/0.6116 & 0.5894/0.5900 & 0.5742/0.5748 & 0.5575/0.5570
& 0.6254/0.6253 \\
\rowcolor{ourgray}
\quad + SIEVE
& 0.7270/0.7273 & 0.6990/0.6987 & 0.6808/0.6800 & 0.6576/0.6591
& 0.6318/0.6333 & 0.6131/0.6149 & 0.5890/0.5891 & 0.5756/0.5759
& 0.6467/0.6473 \\
\midrule
MoMKE
& 0.7760/0.7742 & 0.7512/0.7488 & 0.7362/0.7344 & 0.7147/0.7114
& 0.6954/0.6918 & 0.6840/0.6826 & 0.6651/0.6619 & 0.6437/0.6405
& 0.7083/0.7057 \\
\rowcolor{ourgray}
\quad + SIEVE
& 0.7925/0.7908 & 0.7682/0.7659 & 0.7478/0.7442 & 0.7252/0.7220
& 0.7108/0.7087 & 0.6914/0.6885 & 0.6805/0.6773 & 0.6555/0.6545
& 0.7215/0.7190 \\
\bottomrule
\end{tabular}%
}

\end{table*}

\end{document}